# Next Token Prediction Is a Dead End for Creativity: Why It's Impossible to Lose Yourself in the Moment

Ìbùkún Ọlátúnjí[1]     Mark Sheppard[2]*

[1]Computational Foundry, Swansea University, Crymlyn Burrows, Skewen, Swansea SA10 6JW, UK
[2]University of Kent, Canterbury, Kent CT2 7NZ, U

## Abstract

This position paper argues that token prediction is fundamentally misaligned with real creativity. While next-token models have enabled impressive advances in language generation, their architecture favours surface-level coherence over spontaneity, originality, and improvisational risk. In contrast, creative acts, particularly in live performance domains, require dynamic responsiveness and stylistic divergence, enabling humans to transcend pre-learned patterns in the moment. We use battle rap as a case study to expose the limitations of predictive systems, demonstrating that they cannot truly engage in adversarial or emotionally resonant exchanges. As a result, such models fail to support the interactive flow states where human creators "lose themselves in the moment." Rather than pursuing greater predictive accuracy, we argue that AI research should embrace dialogue as a form of co-negotiated creative agency. This shift calls for approaches that prioritize real-time interaction, rhythmic alignment, and adaptive generative control. By reframing creativity as an interactive process rather than a predictive output, we offer a vision for AI systems that are more expressive, responsive, and aligned with human creative practice.

## Background

Large Language Language Models (LLMs) have revolutionised the landscape of text generation ((Brown et al. 2020; Radford et al. 2019). A recent study found GPT-4 outperformed 100,000 humans on the Divergent Association Task (DAT) (Olson et al. 2021), indicating strong semantic creativity (Bellemare-Pepin et al. 2024). LLMs have also found success in coding (Starace et al. 2025), games (Silver et al. 2018), and music generation (Copet et al. 2024). However, their dominance masks critical limitations that (i) the world can be substantially represented through language and (ii) that language can be modelled as a unidirectional sequence of tokens (Vaswani et al. 2017; Patel 2023). The argument for this is that a single unidirectional language model can produce complex, structured outputs (e,g., code, poetry) when guided, without task-specific architectures. The versatility across modalities and formats is taken as evidence that language (and related symbol sequences) can be sufficiently represented as one-directional token streams Patel 2023). However, despite LLMs successes, this assumption is misaligned with the demands of real-time, interactive communication. This is particularly the case in high-stakes, performative contexts like debating (Official 2023),negotiation (Bell and Valley 2020), or freestyle rap battles(Ọlátúnjí et al. 2025). We use battle rap as a case study that foregrounds the limitations of token-level prediction. Unlike traditional dialogue generation tasks, battle rap involves spontaneous counterpoint, rapid adaptation to an adversary's style and substance (Caffeine 2020), and live calibration to audience reaction and rhythm (Santos 2023). Success depends on delivering the most contextually disarming, rhythmically inventive, and stylistically appropriate response (Flip Top Battles 2025). In this paper we position battle rap as a test bed for human-machine co-performance. By examining the genre through a computational lens, we aim to motivate the development of new technical approaches and evaluation standards that reflect the dialogic, embodied, and adversarial nature of language in the wild.

| Capability | Next-Token Prediction Models | Improvisational Co-Creation |
|---|---|---|
| Temporal Strategy | Short-term, reactive continuation | Strategic planning across turns |
| Context Handling | Fragile under topic or tone shifts | Resilient to abrupt redirection |
| Creativity Mode | Coherence-driven generation | Divergence and surprise prioritized |
| Adversarial Engagement | No competitive modelling | Core to interaction and impact |
| Dialogue Awareness | Isolated outputs; no anticipation | Opponent-aware, turn-sensitive |

Table 1: Comparison of Next-Token Prediction Models and Improvisational Co-Creation

Corresponding authors: 2030349@swansea.ac.uk, ms2403@kent.ac.uk

Table1 compares the functional capabilities of next-token prediction models with the demands of improvisational co-creation in contexts like rap battles. While next-token models operate reactively, optimizing for local coherence, improvisational performance requires proactive planning across multiple turns. These systems often struggle with abrupt shifts in context, whereas skilled human performers adapt fluidly to changes in tone, topic, or audience. Creativity in language models tends toward plausible continuation, whereas improvisation thrives on divergence and surprise. Furthermore, adversarial reasoning and dialogic awareness, central to live, competitive co-creation, are largely absent in current autoregressive systems. The comparison highlights key gaps between conventional language modelling and the complexity of human improvisation.

## Introduction

This paper argues that *token prediction is a dead end for creativity,* particularly in modelling real-time, interactive, and performative forms of human expression. While large language models (LLMs) have achieved remarkable fluency in static settings, their performance falters in live, creative exchanges where timing, responsiveness, and emotional context are essential. Nowhere is this failure more visible than in battle rap, an art form defined by call-and-response, adversarial tension, and spontaneous lyrical invention. Token prediction optimizes for plausible sentence continuation, but not for turn-taking, anticipation, or rebuttal—core elements of human dialogue and performance. In battle rap, the human MC is engaged in a strategic exchange that values not just linguistic correctness, but performative style, emotional delivery, rhythm, and cultural reference. These requirements stretch beyond the capabilities of traditional autoregressive models. We argue that the future of real-time human-AI interaction lies not in scaling token prediction, but in rethinking interaction as a dynamic, multimodal game. Battle rap, as both a test bed and metaphor, makes clear the need for adversarial feedback, prosodic alignment, and co-creative performance. This paper positions battle rap as a lens through which to critique current architectures, and as a foundation for building more responsive, human-centred AI.

## Is Creativity "Only Human"?

Human creativity has been described as being 'a mysterious and complex phenomenon' (Boden 2009; Zhang, Sjoerds, and Hommel 2020). Nevertheless, a body of research describes the key concepts surrounding creativity. These provide a foundation for working definitions and models (Boden 2009; Runco 2010; Cropley 2006; Csikszentmihalyi 1997), and Boden argues that creativity is 'the ability to come up with ideas or artefacts that are new, surprising, and valuable. It enters virtually every aspect of life as an aspect of human intelligence in general.' Accordingly, creativity is grounded in everyday abilities (e.g., conceptual thinking, perception, and memory) and everyone is creative, to a varying degree. It is an almost universal feature of humanness as opposed to a quality that only some humans have. Value is perhaps the most fluid of the dimensions of creativity in Boden's model. It is necessary that some value is attributed to an idea for it to be creative. It is not clear however what kind of value is attributable: Social, cultural, economic, or some other measure. Boden states that values are "difficult to recognize, more difficult to put into words, and even more difficult to state really clearly because creativity by definition involves not only novelty but value, and because values are highly variable, it follows that many arguments about creativity are rooted in disagreements about value.' (Boden 2009).

From a cognitive perspective, creativity can be seen as comprising two measurable ingredients that are crucial to the creative processes: divergent and convergent thinking. Divergent thinking represents a style of thinking that allows idea generation, in a context where the selection criteria are relatively vague and more than one solution is correct. Divergent thinking can be characterised by flexibility of the mind (Runco 2010; Runco and Acar 2012; Lewis and Lovatt 2013). Guilford's Alternative Uses Task (AUT) is a commonly used measure of divergent thinking ability: participants are presented with familiar objects (e.g., a brick) and they have to generate as many different uses of the object as possible (Guilford et al. 1960). In contrast, convergent thinking is a process of generating one possible solution to a particular, well-defined problem (Zhang, Sjoerds, and Hommel 2020; Cropley 2006). Mednick's Remote Associates Test (RAT) is commonly used in convergent thinking literature (Mednick 1968). Within the RAT, participants are presented with three unrelated words (e.g., cocktail, dress, and birthday) and identify the common associate ("party"). The total number of correct answers is used to evaluate their level of convergent thinking. Success within the task requires tight top-down constraints, as there is only one possible answer per item. Cognitively, this suggests a control state with a strong goal-directed bias toward persistence (Hommel 2015). It is suggested therefore, that divergent and convergent thinking are likely to serve different purposes, and to satisfy different task demands (Runco and Acar 2012; Hommel 2015).

The AUT and RAT do not represent pure measures of divergent and convergent thinking or their underlying processes. In both tasks, participants are required to hold one goal-related concept while switching between other possible or actual alternatives. It is likely therefore that the tasks require divergent and convergent thinking, and both flexibility and persistence (Bernard A. Nijstad and Baas 2010). The Dual Pathway to Creativity (DPC) Model distinguishes between a flexibility route and a persistence route to creative performance, and it assumes that creative outputs (called *products* within the literature) rely on the two routes to different degrees. The focus of attention varies as a function of task demands. When the task ambiguity is high, attention is defocused, resulting in slower processing on the task. In contrast, when the task ambiguity is low, attention is focused, resulting in faster processing on the task (Vartanian 2009). To create highly original ideas, flexibility is needed to switch between routes.

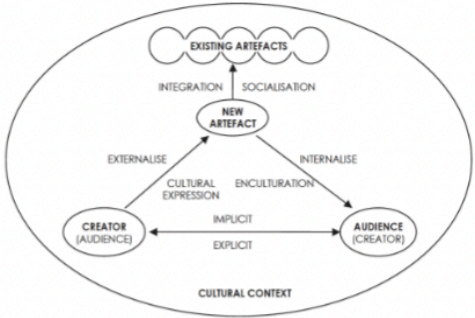

Figure 1: Creativity as a Social System.

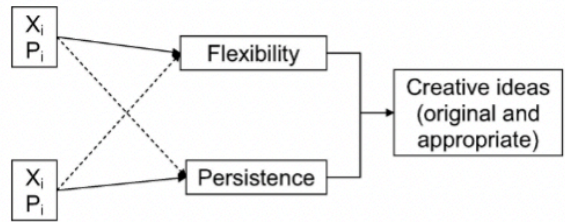

Figure 2: The dual pathway to creativity model (Bernard A. Nijstad and Baas 2010). Flexibility and persistence are shaped by situational (Xi) and dispositional (Pi) factors, with differing strengths indicated by solid vs. dotted lines.

In recent decades, creative cognition has shifted from a person-centred perspective to a socially dynamic one (John-Steiner 1992). Within this framework, socio-cultural theories of creativity and learning have gained recognition, emphasizing that 'culture clearly has a profound influence on the conceptualisation of creativity and on creative expression' (Gong et al. 2023). Creativity is socio-cultural because: a) the set of skills and types of knowledge that individual actors possess are developed through social interaction; b) creativity in itself is often the result of explicit moments of collaboration between individuals; c) creativity is largely defined by social judgement or validation; and d) creativity exists only in relation to an established ensemble of cultural norms and products that both aliment the creative process and integrate its outcomes.

Figure 1 and Figure 2 illustrate two complementary perspectives on human creativity. The first frames creativity as a social system, highlighting how cultural, institutional, and interpersonal dynamics shape creative production and recognition. The second, the Dual Pathway to Creativity model emphasizes individual cognitive processes, identifying two distinct routes to creative output: a flexibility and a persistence pathway. Both pathways are influenced by situational (denoted with Xi) and dispositional (denoted with Pi) factors. However, some situational and dispositional variables affect the flexibility pathway more strongly than the persistence pathway, and vice versa, which is indicated by solid (stronger relation) and dotted (weaker or negative relation) lines, respectively. Together, these models underscore that creativity emerges not only from internal traits or strategies, but also from social and situational contexts, an important contrast to prevailing assumptions in current machine creativity systems.

**Battle Rap as a Creativity Test**

Due to its structured, yet open-ended nature, battle rap presents a unique opportunity to explore human vs machine creation beyond next-token prediction. Unlike conventional dialogue systems, battle rap demands real-time adaptation, adversarial reasoning, and dynamic control over linguistic and rhythmic structures. Participants (known as MCs) must not only generate fluent responses but also anticipate and strategically counter an opponent's lines, while maintaining stylistic coherence and crowd engagement. As such, battle rap constitutes a high-stakes improvisational domain that is both computationally meaningful and technically demanding. Its combination of constraints and creative freedom offers a rigorous benchmark for testing AI models' abilities in real-time language generation, improvisation, and multimodal fluency. Addressing this domain effectively requires novel methods that move toward active, context-sensitive generation and interaction.

| Component | Linguistic Features | Musical Features |
|---|---|---|
| Content | Storytelling, metaphor, wordplay | Phonetic manipulation, assonance, consonance |
| Flow | Rhyme schemes, syntactic complexity | Rhythm, tempo, syncopation |
| Delivery | Prosody, emphasis, vocal texture | Timbre, breath control, cadence |

Table 2: Key Elements of Rap

Table 2 outlines the core components of rap performance, distinguishing between their linguistic and musical dimensions. While rap is often analysed through its verbal content, such as metaphor, rhyme, and syntactic structure, it also relies heavily on musical features like rhythm, cadence, and vocal delivery. This dual framing highlights rap as a multimodal art form that blends linguistic complexity with performative musicality. As a creative act, writing rap verse requires an extensive vocabulary, mastery of complex rhyme patterns, and broad subject knowledge across diverse topics (Bradley 2017; Liu et al. 2012). An fMRI study of freestyle rap points to the requirement for rapid linguistic processing, as MCs must generate meaningful, rhyming phrases in real-time while adhering to tempo and rhythm constraints (Liu et al. 2012). The process shares characteristics with the flow state, where cognitive effort, heightened focus, and automaticity merge to enable fluid creative expression (Csikszentmihalyi 1997).

## Towards Real-Time Improvisation

The proposed AI-driven rap system (Figure 3) integrates voice cloning, cadence modelling, and rhyme structuring to generate adaptive lyrical responses. The MC's speech acts as both input and feedback, guiding generation. A voice clone synthesizes replies in the MC's voice and style for natural delivery. The *current style module* evolves through improvisation and an *archive corpus* of past performances. A cadence algorithm analyses syllabic timing and rhythm, while a rhyme module scores rhyme density and complexity. These components interact dynamically, with archival data reinforcing stylistic coherence. A feedback loop enables continuous refinement, allowing the AI to adjust based on its own output and the MC's voice. Merging historical influence with real-time adaptation, the system produces increasingly precise, context-aware responses that emulate human performance in rap battles.

Figure 3: Real-Time Improvisation System

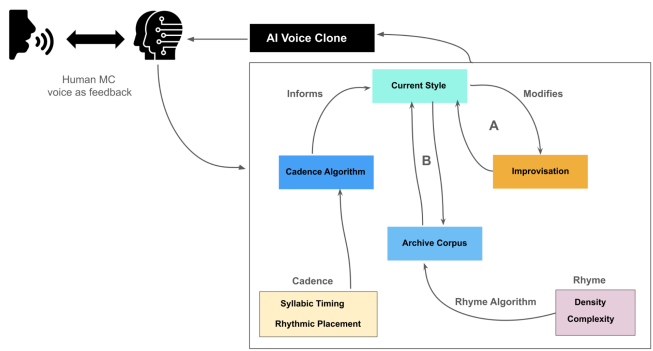

Cadence, the rhythmic structure of syllables and stress, is essential to rap delivery. A simplified description of flow[1] is that it combines cadence with rhyme and timing to form a rapper's stylistic fingerprint. Following Adams (Adams 2009) we define flow as 'all of the rhythmical and articulative features of a rapper's delivery of the lyrics.' In contrast to this, most LLMs optimise for semantic coherence and are unable to model rhythmic constraints or beat alignment (Ọlátúnjí et al. 2025). To capture cadence, we use algorithms to segment input by syllable, track stress patterns, and align output with any musical timing. Real-time cadence modelling supports not only delivery alignment but also generation of rhythmically coherent and stylistically adaptive verses, a prerequisite for believable human-AI rap co-performance. Table 3 outlines the key computational demands of live battle rap and contrasts them with the limitations of current next-token prediction models. It highlights five core domains where human performance expectation, such as real-time rhythmic response, adversarial reasoning, spontaneous wordplay, and adaptive delivery pose challenges for autoregressive language models. While human MCs can dynamically adjust to beat, opponent, and audience, next-token systems lack temporal precision, strategic awareness, and multimodal fluency. This comparison illustrates the gap between human improvisational creativity and the sequential, text-only nature of current generative AI models.

**Modelling Human Responses: The Ghost in the Machine**

Artificial intelligence systems are in constant development to mimic human thought processes and creativity. Creativity in humans, however, is intrinsically linked to subjective experience, which involves a combination of intuition, experience-based intentionality, and spontaneity to produce ephemeral situational responses. This is frequently derived from a variety of indirect approaches, such as distraction techniques and reflective incubation periods (Vernon, Hocking, and Farahar 2016; and 2024; Marrone, Cropley, and and 2024; Ritter and Dijksterhuis 2014); creating a complex, multimodal environment wherein inspiration can take place (Vernon, Hocking, and Farahar 2016; Hermawan et al. 2023). In contrast, it can be argued that AI systems are ultimately reductive, simplifying complex human behaviours into mechanistic frameworks (Riedl 2014; Fowler 2022). The AI can learn specific tasks, but there is a focus on predictive accuracy at the expense of model complexity, resulting in over-fitting by learning non-generalistic features (Pezzulo et al. 2024).

---

[1] Here, flow refers to rhythmic or lyrical delivery, not the psychological concept by Csikszentmihalyi, which describes a state of deep, immersive focus.

| Challenge Domain | Battle Rap Requirement | Limitation of Next-Token Models |
|---|---|---|
| Temporal Dynamics | Real-time response within tight rhythmic constraints | Slow or off-beat generation; latency issues |
| Adversarial Reasoning | Counter opponent's intent and tone mid-performance | No awareness of opponent's strategy |
| Improvisational Creativity | Spontaneous wordplay and flow innovation | Emphasis on coherence over divergence |
| Multimodal Fluency | Synchronized lyric delivery over beats | No native modeling of rhythm or audio constraints |
| Audience Adaptation | Adjust style/tone based on crowd reaction | Lacks real-time feedback integration |

Table 3: Computational Demands of Battle Rap vs. Capabilities of Next-Token Prediction

This removes subjective intentionality and the concept of consciousness from the creative process, removing critical dimensionality, and limiting the potential to generate truly spontaneous outcomes..

Modelling human subconscious processing as a generative AI paradigm would be the ultimate development goal. Currently, even the best AI systems have proven only to be convincing creative mimics; they lack the granular inputs that a human possesses. The goal of generative AI development should be to enhance human creativity using intuitive and expansive AI tools rather than replace it (Pescapè 2024). Conversely, the elastic concepts of authenticity, emotional resonance, and spontaneity can only be informed by human input and artistic evolution (Bringsjord, Bello, and Ferrucci 2003). Thus, creating AI is akin to disembodiment; and this divergence from the human experience will only be perpetuated with each future iteration (Wang and Baker 2024). "It could be argued that generative AI is one of the most beautiful and important inventions of the century – a 21st-century 'mirror' in which we can see ourselves in a new and revealing light. However, when we look behind the mirror, there is nobody there." (Pezzulo et al. 2024)

**Artificial vs *Real* Creativity**

Runco (Runco 2023) argues that while AI can produce outputs that appear original and effective, thus satisfying surface-level definitions of creativity, it fundamentally lacks the attributes that render creativity authentic namely, intrinsic motivation, intentionality, and self-aware agency. Without the capacity for conscious choice or internal drive, AI cannot meaningfully originate ideas. Its output, therefore, is more accurately described as *artificial creativity* or *pseudo-creativity:* a simulation of creative products that lacks the generative processes and subjective experience underpinning genuine human creativity. We build on Runco's critique by extending the concept of authenticity beyond the individual. Rather than viewing creativity as the exclusive domain of the human, we propose that creative authenticity may emerge through a co-constitutive relationship between the human and their personalised AI. This reframing positions authenticity not as an inherent human trait, but as a function of intentional, expressive alignment between user and system. This view aligns with geographic perspectives that conceptualise creativity as a relational and *more-than-human* process. Lundman and Nordström (Lundman and Nordström 2023) describe "co-creative spatiality," where human and non-human agencies, including AI, intertwine within spatial, material, and symbolic contexts. Similarly, work in computational creativity argues for recognising AI systems as collaborators within hybrid creative ecosystems (Colton and Wiggins 2012; Davis 2021). Rather than treating AI as a failed attempt at human creativity, we suggest that authenticity itself may increasingly be co-authored, emerging from a system that includes both the human and their deeply personalised AI. The critiques of AI creativity suggest that enhancing output quality alone is insufficient. In domains like battle rap, creative authenticity cannot be reverse-engineered from plausibility. Instead, we shift focus to how co-creative systems might support interactive expressiveness and real-time responsiveness. Table 4 outlines computational approaches that extend beyond token prediction to support dynamic, situated co-creation. These techniques, ranging from adversarial learning to cadence modelling, represent a reorientation of system design from static generators towards improvisational partners. One such strategy, Creative Beam Search (CBS) (Franceschelli and Musolesi 2024a), exemplifies how structured sampling and internal evaluation mechanisms can simulate the generative and reflective phases of human creativity. By combining Diverse Beam Search (Kasai et al. 2024) with LLM-as-a-Judge self-evaluation (Zheng et al. 2023), CBS approximates a simple yet effective generate-and-test loop that may be adapted for interactive domains like freestyle rap. These strategies represent a shift from purely predictive models toward architectures capable of dynamic, multimodal co-creation.

| Method | Capabilities | Relevance to Rap Improvisation |
|---|---|---|
| Adversarial Imitation Learning (e.g., GAIL) | Learns policies from expert behaviour via adversarial training | Enables AI to mimic MC tactics, generate rebuttals, and refine strategy in real time |
| Transformer-based Cadence Modelling | Captures rhythmic and prosodic constraints in language | Aligns generated lyrics with beats; preserves flow and timing |
| Human-in-the-Loop (HITL) Training | Incorporates human feedback during generation | Supports collaborative co-creation; preserves human style and intent |
| Speech-to-Text and Beat Tracking Integration | Real-time alignment of lyrics to vocal input and rhythm | Allows seamless call-and-response interaction between MC and AI |

Table 4: Computational Approaches to Real-Time Rap Co-Creation

## Alternative Perspectives

The next-token prediction architectures underpinning large language models (LLMs) have demonstrated significant potential for real-time creative interaction. Modern LLMs can produce outputs in domains like storytelling (Riedl 2021), poetry (Sawicki et al. 2023), and rap lyrics (Xue et al. 2021) with at high quality (Franceschelli and Musolesi 2024b), challenging the notion that they merely regurgitate training data. In creative writing tasks, for example, these models can generate original work that surprises human users; in many cases human-users are not able to distinguish between AI-authored and human-authored texts (Porter and Machery 2024; Karsdorp, Manjavacas, and Kestemont 2019; Köbis and Mossink 2021) Next-token predictors are adaptive and evolving in their creative capabilities. Techniques like prompt iteration and ensemble querying can elicit a diversity of ideas analogous to a human brainstorming session. A recent study showed that when an LLM is queried multiple times for a task, its "collective creativity" can rival that of a group of 8–10 humans (Newkeen 2024). By requesting numerous responses and pooling the results, LLMs were able to compete with small human teams on divergent-thinking and problem-solving challenges (Newkeen 2024) With the correct interaction strategies, next-token models exhibit a breadth of ideas and inventiveness comparable to, and often surpassing, human collaborators.

Empirical research indicates that LLMs not only produce creative artifacts, but may do so using strategies reminiscent of human creativity. For instance, Nath et al. (Nath, Dayan, and Stevenson 2024; Max Planck Institute 2025) found that LLMs tackle creative tasks (e.g., inventing new uses for everyday objects in the Alternative Uses Test) in similar ways to humans, employing flexible and persistent approaches (Bernard A. Nijstad and Baas 2010). Such parallels hint that next-token models can emulate cognitive patterns of creativity such as those in Figure 2. As model architectures and training techniques advance we might expect even richer real-time interactions. For example, an AI MC might build on its opponent's last verse with wit and coherence in real time.[2] However, from an interactional standpoint, systems can be designed to emulate the external signs of losing oneself through delay, stylistic deviation, or sudden creative leaps. These behaviours might be perceived as flow-like by human interlocutors. However, that perception alone is insufficient for authenticity. True co-creativity requires not just plausible output, but the intentional negotiation of meaning, context, and relational timing, capacities that token prediction, as a unidirectional paradigm, cannot account for on its own. The counterargument also questions the assumption that AI-generated content is inherently inauthentic or lacks creative agency. Detractors often posit that true creativity requires human-like intentionality or intrinsic motivation, a *soul* behind the art.

Not all definitions of creativity demand an internal intent. Creativity is commonly defined by the novelty and value of the outputs (Newkeen 2024), regardless of how or why it was produced. Some scholars argue that intrinsic motivation is not a strict requirement for creativity, except perhaps for a special exemplary form of it (Paul and Stokes 2023). An AI system might lack the spontaneous urges of a human artist, but it can still generate outcomes that meet creative criteria. If an LLM produces a battle-rap verse that is linguistically original, surprising, and apt to the challenge, that verse is creative in its own right if it provokes engagement and admiration much as a human's would. Also, the absence of a singular personal authorship in AI-generated work does not preclude creativity; one can view the AI, its training data, and the user as a new kind of distributed creative agent.

---

[2] Most rap battles involve pre-written verses delivered with improvised timing or emphasis (e.g., adapting to a beat or crowd reaction). Fully improvised freestyling, known as "off-the-dome" is considerably harder and less common in formal competitions.

What matters for authenticity in practice is often the impact of the work on its audience, not the presence of a conscious originator. By this measure, dismissing AI outputs as "mere imitation" overlooks the ways in which LLMs contribute original expressions and ideas. They frequently appear creative to observers (Franceschelli and Musolesi 2024b) and in collaborative settings they can inspire and augment human creativity rather than simply aping it (Arnold, Volzer, and Madrid 2021). A recent argument has even been made that are on the verge of a AI revolution termed *Software with a Soul*. The originators of the term define it as '" artificial intelligence that is capable of *humanness*…stochastic creativity, empathy, intuition, ability for connection and even emotional expression." (Piñol 2024) [emphasis added].

## Immersion in Human–AI Systems

*Losing oneself* in the creative act suggests immersion in a psychological state of flow, marked by spontaneity, deep absorption, and a blurring of self–other boundaries (Csikszentmihalyi 1997). Such states often emerge from perceived lack of control (Mulatti and Treccani 2023) or non goal-directed cognition such as mind-wandering (Gable, Hopper, and Schooler 2019). Crucially, such states depend on agency (the capacity for self-directed action), embodiment (a physically situated perspective), and subjective temporality (a lived experience of time).

Current AI systems lack these prerequisites. Without intrinsic motivation, self-awareness, or temporal experience beyond left-to-right token prediction, an LLM cannot meaningfully "lose itself." There is no self to dissolve, no lived context to disorient, no inward transformation. As Graves wrote, "Only a poet of experience can hope to put himself in the shoes of his predecessors... and judge their poems by recreating technical and emotional dilemmas which they faced while at work on them" (Schmidt 1999, p. 7). Despite stylistic fine-tuning, reinforcement learning, or sampling techniques like Creative Beam Search, LLMs produce output through deterministic processes. What appears as spontaneity is a simulation, lacking the interiority and experiential grounding it seems to express.

### Helping Humans Lose Themselves

While current models may not experience immersion themselves, a more constructive line of inquiry asks how AI systems might facilitate immersive states in humans. In freestyle rap battles, for instance, flow is not merely a matter of linguistic coherence, it involves rapid adaptation to adversarial input, rhythmic synchronisation with a beat, and the spontaneous expression of wit, emotion, and identity. These performances unfold under high-tempo, high-stakes conditions where responsiveness and timing are essential.

In rap battle contexts, LLMs do not possess human capabilities, but they can still serve a valuable supporting role. If designed with interaction-awareness, real-time inference, and multimodal feedback mechanisms, such systems could reduce friction in human expression, adapt to creative cues in-situ, and scaffold improvisation in ways that augment, rather than disrupt, creative flow.

Flow should not be understood as an internal state of the AI, but as an emergent property of the human-machine interaction. It reflects how effectively the system amplifies human spontaneity, minimizes latency, and enables real-time expressivity. The aim is not for the AI to experience immersion, but to facilitate it: by staying contextually aligned, reacting dynamically, and preserving the unpredictability that characterises human creativity. This reframes the role of AI in creative domains, not as an originator of conscious experience, but as a co-creative infrastructure designed to support and enhance the human performer's ability to "lose themselves" in the moment.

## Conclusion

This paper has argued that token prediction is insufficient for modelling creativity, particularly in real-time, interactive, and performative domains. Using freestyle rap battles as a case study, we highlighted the structural limitations of autoregressive language models in capturing spontaneity, adversarial engagement, and rhythmic improvisation. Unlike static text generation tasks, battle rap demands dynamic responsiveness, stylistic adaptation, and strategic timing, requirements that expose the expressive and architectural gaps of current LLMs.

Far from being a niche example, freestyle rap serves as a stress test for broader challenges in conversational AI. Its high-intensity constraints compress many fundamental issues in human-machine interaction—timing, rebuttal, cadence, and performance—into a vivid, evaluable format. These dynamics reveal why semantic plausibility alone is not enough; co-creative agents must engage with context in real time and under pressure. If creativity is the goal, generative systems must evolve from token-wise prediction to architectures capable of embodied, adaptive, and goal-directed interaction. This includes planning, multimodal perception, and turn-taking grounded in real-time feedback loops. As Csikszentmihalyi (Csikszentmihalyi 1997) and Runco (Runco 2023) state,, creativity is more than output, it is a process tied to immersion, intent, and dynamic co-regulation. These remain outside the reach of current predictive models.

To advance this agenda, we put forward the following research challenges for the machine learning community:

• Develop interaction-first training paradigms that support collaborative improvisation and adaptive turn-taking.
• Create evaluation benchmarks for real-time co-creative systems, incorporating multimodal alignment, cadence, and adversarial responsiveness.
• Design model architectures that integrate planning, rhythm-awareness, and feedback sensitivity into generative loops.
• Explore hybrid approaches that combine symbolic reasoning, reinforcement learning, and user-in-the-loop optimisation for performative AI.

We invite the machine learning community to reimagine dialogue as improvisation rather than completion; to see language as co-authored action, not just statistically plausible output. Whether in education, negotiation, or the arts, future AI must support collaborative immersion. Freestyle rap highlights what is missing and what might yet be built, if we shift our focus from static benchmarks to dynamic interaction.

## Acknowledgements

This work was supported by the Engineering and Physical Sciences Research Council.